\documentclass[preprint,11pt]{elsarticle}
\usepackage[utf8]{inputenc} 
\usepackage{float}
\usepackage{afterpage}
\usepackage{graphicx} 
\usepackage{multirow}
\usepackage[final]{changes}
\usepackage{booktabs} 
\usepackage{array} 
\usepackage{paralist} 
\usepackage{verbatim} 
\usepackage{tikz}
\usepackage{diagbox}
\usepackage{mathtools}
\usepackage{algorithm}
\usepackage{algorithmicx}
\usepackage{algpseudocode}
\usetikzlibrary{arrows,shapes,chains}
\usepackage{braket,amsfonts}
\usepackage{amsmath,dsfont}
\usepackage{harpoon}
\usepackage{amssymb}
\usepackage{mathrsfs}
\usepackage{graphics}
\usepackage{color}
\usepackage{epsfig}
\usepackage{bbm}
\usepackage{xfrac}
\usepackage{fancyhdr}
\usepackage{makecell}
\usepackage{enumerate}
\usepackage{bm}
\usepackage{xcolor}
\usepackage{setspace}
\usepackage{geometry}
\def\dd{\mbox{d}}

\usepackage{amsthm}

\usepackage[caption=false]{subfig}

\usepackage{pgfplots}

\newtheorem{example}{\textup{\textbf{Example}}}

\usepackage{amssymb}

\journal{Journal of Computational Physics}

\begin{document}

\begin{frontmatter}



\title{A Spectral Approach for Learning Spatiotemporal Neural Differential Equations}


\author[NYU]{Mingtao Xia}
\ead{xiamingtao@nyu.edu}

\author[UCLA]{Xiangting Li}
\ead{xiangting.li@ucla.edu}

\author[OX]{Qijing Shen}
\ead{qijing.shen@ndm.ox.ac.uk}

\author[UCLA]{Tom Chou\corref{cor1}}
\ead{tomchou@ucla.edu}

\cortext[cor1]{Corresponding author}

\address[NYU]{Courant Institute of Mathematical Sciences, New York University, New York, NY
  10012, USA}
\address[UCLA]{Department of Computational Medicine, UCLA, Los Angeles, CA
  90095, USA}
\address[OX]{Nuffield Department of Medicine,
 Oxford University, Oxford OX2 6HW, UK}


\begin{abstract}
Rapidly developing machine learning methods has stimulated research
interest in computationally reconstructing differential equations
(DEs) from observational data which may provide additional insight into
underlying causative mechanisms.  In this paper, we propose a novel
neural-ODE based method that uses \textit{spectral expansions} in
space to learn spatiotemporal DEs. The major advantage of our spectral
neural DE learning approach is that it does not rely on spatial
discretization, thus allowing the target spatiotemporal equations to
contain long-range, nonlocal spatial interactions that act on
\textit{unbounded} spatial domains. Our spectral approach is shown to
be as accurate as some of the latest machine learning approaches for
learning PDEs operating on bounded domains.  By developing a spectral
framework for learning both PDEs and integro-differential equations,
we extend machine learning methods to apply to unbounded DEs and a
larger class of problems.
\end{abstract}



\end{frontmatter}

\section{Introduction}
There has been much recent interest in developing
machine-learning-based methods for learning the underlying physics
from data. Although machine learning approaches have been proposed for
many types of inverse problems
\citep{bar2019unsupervised,hsieh2018learning,ruthotto2020deep,sirignano2020dpm},
most of them make prior assumptions on the specific form of the
underlying partial differential equation (PDE) and use discretization
\textit{i.e.}, grids or meshes, of a bounded spatial variable $x$ to
approximate solutions to a presupposed PDE.  To our knowledge, there
are no methods for learning, without prior assumptions, the
``dynamics'' $F[u; x, t]$ of more general spatiotemporal differential
equations (DEs) such as
\begin{equation}
\partial_t u = F[u; x, t],\quad   x\in\Omega, t\in[0, T].
\label{objective}
\end{equation}
Previous methods either assume a specific form of the operator $F[u;
  x, t]$ in order to satisfy \textit{e.g.}, a conservation law
\citep{brandstetter2021message}, or circumvent learning $F[u; x, t]$
by simply reconstructing the map from the initial condition to the
solution at a later time. Moreover, since most prevailing numerical
methods for time-dependent DEs rely on spatial discretization, they
cannot be applied to problems defined on an unbounded domain
\citep{xia2021efficient}.

In this paper, we propose a spectral-based DE learning method that
extracts the unknown dynamics in Eq.~\eqref{objective} by using a
parameterized neural network to express $F[u; x, t]\approx F[u;x, t,
  \Theta]$.  Throughout this paper, the term ``spatiotemporal DE''
refers to a differential equation in the form of
Eq.~\eqref{objective}.  The formal solution $u$ is then represented by
a spectral expansion in space,
\begin{equation}
u(x, t)\approx  u_N(x, t) = \sum_{i=0}^N c_i(t) \phi_i(x),
\label{spectral_approx}
\end{equation}
where $\{\phi_i\}_{i=0}^N$ is a set of appropriate basis functions
that can be defined on bounded or unbounded domains and
$\{c_i\}_{i=0}^N$ are the associated coefficients.

The best choice of basis functions will depend on the spatial domain.
In bounded domains, any set of basis functions in the Jacobi
polynomial family, including Chebyshev and Legendre polynomials,
provides similar performance and convergence rates; for semibounded
domains $\mathbb{R}^+$, generalized Laguerre functions are often used;
for unbounded domains $\mathbb{R}$, generalized Hermite functions are
used if the solution is exponentially decaying at infinity, while
mapped Jacobi functions are used if the solution is algebraically
decaying \citep{burns2020dedalus, Spectral2011}.  By using such
spectral expansions, a numerical scheme for Eq.~\eqref{objective} can
be expressed as ordinary differential equations (ODEs) in the
expansion coefficients $\bm{c}_N(t)\coloneqq (c_0(t),\ldots, c_N(t))$
\begin{equation}
\frac{\dd \bm{c}_N(t)}{\dd t} = \bm{F}(\bm{c}_N;t).
\label{U_eqn}
\end{equation}

There have been a variety of machine-learning-based methods recently
developed for learning PDEs. Long \textit{et
  al}. \citep{long2018pde,long2019pde} applied neural networks to
learning PDEs that contain only constant-coefficient spatial
differential operators. Convolutional neural networks were then used
to reconstruct these constant coefficients. Xu \textit{et
  al}. \citep{xu2019dl} then studied spatiotemporal PDEs of the form
$\partial_{t}u(x,t) = {\bf a}\cdot (1, u, u^2, u_x, u_{xx}, \ldots)$,
where ${\bf a}$ is the to-be-learned row vector of coefficients
associated with each type of potential term in the PDE.  These methods
imposed an additive form for $F[u; x, t]$. A neural PDE solver that
partially relaxes the need for prior assumptions on $F[u; x, t]$ was
proposed in \citep{brandstetter2021message}.  This approach learns the
mapping $\bm{u}^k\rightarrow\bm{u}^{k+1}$ for time-homogeneous
conservation PDEs of the form $\partial_{t} u + \nabla\cdot J(u)
=0$. Since $\bm{u}^{k}$ is the solution on grid points at times
$t_{k}$, this method also relies on spatial discretization and can
only be used to learn bounded-domain local operators.  Additionally, a
Fourier neural operator (FNO) approach \citep{anandkumar2020neural}
which learns the mapping between the function space of the initial
condition $u_0(\cdot, 0)$ and the function space of the solution
within a later time range $u(\cdot, t), t\in[t_1, t_2]$ has also been
developed \citep{li2020fourier}. Generalizations that include basis
functions other than the Fourier series were developed in
\citep{fanaskov2022spectral}. However, such methods treat $x$ and $t$
in the same way using nonadaptive basis functions, which cannot be
efficiently applied to unbounded-domain spatiotemporal problems where
basis functions often need to be dynamically adjusted.

Recently, a spectrally adapted PINN (s-PINN) method was proposed to
solve specified unbounded-domain PDEs~\citep{xia2022spectrally}. The
method expresses the underlying unknown function in terms of spectral
expansions, does not rely on spatial discretization, and can be
applied to unbounded domains. However, like many other approaches, the
s-PINN approach assumes that the PDE takes the specific form
%
$u_t = F(u, u_x, u_{xx},...) + f(x, t)$,
%
where all terms in $F(u, u_x, u_{xx},\ldots)$ are known and only the
$u$-independent source term $f(x, t)$ can be learned because the
neural network only accepts times $t$ as inputs.  Therefore, the
s-PINN method is limited to parameter inference and source
reconstruction.

\begin{figure}[htb]
\begin{center}
\includegraphics[width=6.2in]{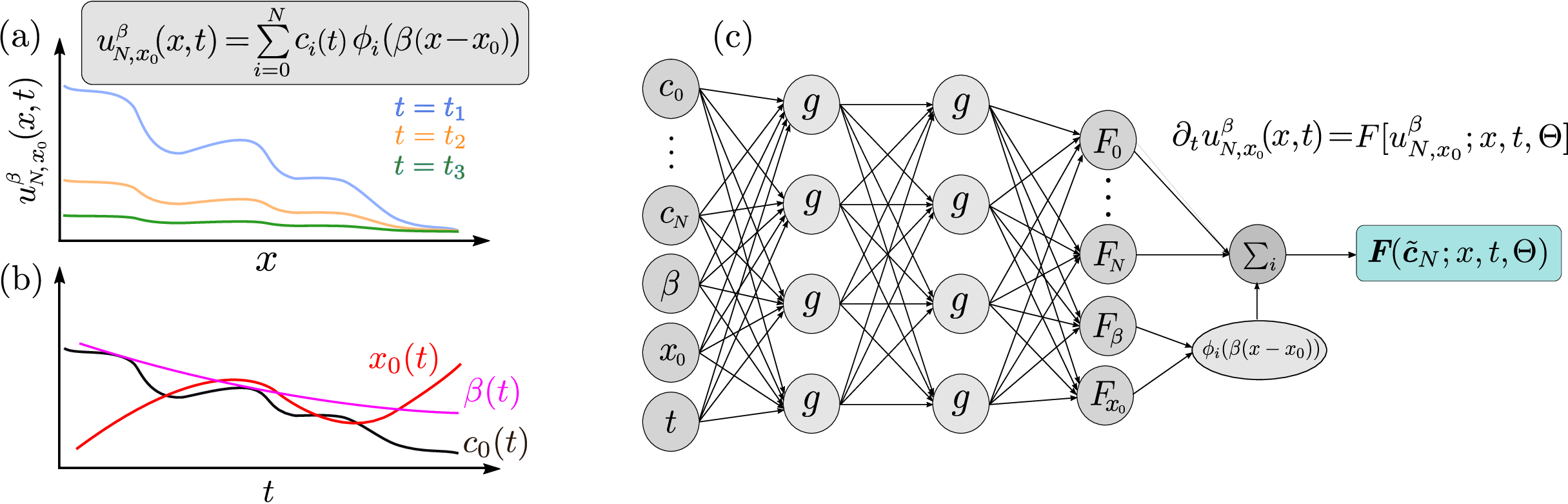}
\caption{\small (a) A 1D example of the spectral expansion in an
  unbounded domain with scaling factor $\beta$ and displacement $x_0$
  (Eq.~\eqref{spectral_unbounded}). (b) The evolution of the
  coefficient $c_0(t)$ and the two tuning parameters $\beta(t),
  x_0(t)$. (c) A schematic of how to reconstruct Eq.~\eqref{objective}
  satisfied by the spectral expansion approximation $u_{N,
    x_0}^{\beta}$.  The time $t$, expansion coefficients $c_i$, and
  tuning variables $\beta(t)$, and $x_0$ are inputs of the neural
  network, which then outputs $\bm{F}(\tilde{\bm{c}}_N;t,\Theta)=(F_0,
  ...,F_N, F_{\beta}, F_{x_0})$. The basis functions
  $\phi_i\big(\beta(t)(x-x_0(t))\big)$ are shaped by the
  time-dependent parameters which obey $\tfrac{\text{d}
    \beta}{\text{d}t} \approx F_{\beta}$ and
  $\tfrac{\text{d}x_0}{\text{d} t}\approx F_{x_0}$.}
\label{icml-historical}
\end{center}
\vspace{-0.15in}
\end{figure}

The spectral neural DE learning method proposed here differs
substantially from the s-PINN framework because it does not make any
assumption on the form of the spatiotemporal DE in
Eq.~\eqref{objective} other than that the RHS $F$ does not contain
time-differentials or time-integrals of $u(x,t)$. It inputs both the
solution $u(x, t)$ (in terms of a spectral expansion) and $t$ into the
neural network and applies a neural ODE model \citep{chen2018neural}
(see Fig.~\ref{icml-historical} (c)). Thus, general DEs such as
Eq.~\eqref{objective} can be learned with little knowledge of the RHS.
To summarize, the proposed method presented in this paper

\begin{enumerate}

\item[(i)] does not require assumptions on the $u$-dependence
of $F$ other than it should not contain any time-derivatives or
time-integrals of $u$.  Both spatiotemporal PDEs and
integro-differential equations can be learned in a unified way.

\item[(ii)] directly learns the dynamics of a spatiotemporal
DE by using a parameterized neural network that can time-extrapolate
the solutions, and 

\item[(iii)] does not rely on spatial discretization and can thus
  learn \textit{unbounded}-domain DEs.  By using adaptive spectral
  representations, our neural DE learning method also learns the
  dynamics of the adaptive parameters adjusting the basis functions.
\end{enumerate}

In the next section, we formulate our spectral neural DE learning
method. In Section~\ref{numerical}, we use our spectral neural DE
method to learn the underlying dynamics of DEs, across both bounded
and unbounded domains, by carrying out numerical experiments. Although
our main focus is to address learning unbounded-domain DEs, we perform
benchmarking comparisons with other recent machine-learning based PDE
learning methods available for \textit{bounded-domain} problems.
Concluding remarks are given in Section~\ref{summary}.  Additional
numerical experiments and results are given in the
Appendix. Below is a table of the notation used in this paper.
\begin{table}[htb]
\caption{{\small\textbf{Overview of variables.} Definitions of
    the main variables and parameters used in this paper.}}
\vspace{2mm}
\centering
\begin{tabular}{c|l}\hline
\textbf{Symbol} & \textbf{Definition} \\
\hline
\,\,\, $N$\,\, & spectral expansion order \\ 
\,\,\, $\Theta$\,\, & neural network hyperparameters \\
\,\,\, $\phi_i$\,\, &  $i^{\rm th}$ order basis function in a spatial spectral expansion\\ 
\,\,\, $\beta(t)$\,\, & spatial scaling factor in basis functions: $\phi_i(\beta(x-x_0))$ \\
\,\,\, $x_0(t)$\,\, & spatial translation in basis functions:
$\phi_i(\beta(x-x_0))$ \\
\,\,\, $u_{N, x_0}^{\beta}(x, t)$\,\, & order $N$ spectral expansion
approximation: \\ &
$u_{N, x_0}^{\beta}=\sum_{i=0}^N c_{i}^{\beta(t)}(t)\phi_i(\beta(x-x_0(t)))$ \\
\,\,\, $\hat{\mathcal{H}}_i$\,\, & generalized Hermite function of order $i$ \\
\bottomrule
\end{tabular}
\end{table}



\section{Spectral neural DE learning method}
\label{spectral_neural}
We now develop a spectral neural DE learning method for general
spatiotemporal DEs of the general structure of Eq.~\eqref{objective},
assuming the operator $F[u; x, t]$ does not involve
time-differentiation or time-integration of $u(x,t)$.  However, unlike
in \citep{brandstetter2021message}, the operator $F[u; x, t]$ can take
any other form including differentiation in space, spatial
convolution, and nonlinear terms.

First, consider a bounded spatial domain $\Omega$.  Upon choosing
proper orthogonal basis functions $\{\phi_i(x)\}_{i=0}^N$, we can
approximate $u(x,t)$ by the spectral expansion in
Eq.~\eqref{spectral_approx} and obtain ordinary equations of the
coefficients $\bm{c}_N(t)\coloneqq (c_0(t),...,c_N(t))$ as
Eq.~\eqref{U_eqn}. We aim to reconstruct the dynamics $F(\bm{c}_N; t)$
in Eq.~\eqref{U_eqn} by using a neural network
\begin{equation}
\bm{F}(\bm{c}_N; t)\approx \bm{F}(\bm{c}_N; t, \Theta),
\label{dynamics}
\end{equation}
where $\Theta$ is the set of parameters in the neural network.
We can then construct the RHS of Eq.~\eqref{objective} using $F[u;x,t,
  \Theta]\approx \sum_{i=0}^N F_i(\bm{c}_N; t,\Theta)\phi_i(x)$ where
$F_i$ is the $i^{\text{th}}$ component of the vector $\bm{F}(\bm{c}_N;
t,\Theta)$. We shall use the neural ODE to learn the dynamics
$\bm{F}(\bm{c}_N;t, \Theta)$ by minimizing the mean loss function
$L(u_N(x, t;\Theta), u(x, t))$ between the numerical solution $u_N(x,
t;\Theta)$ and the observations $u(x, t)$.  When data are provided at
discrete time points $t_j$, we need to minimize
\begin{equation}
\sum_{m=1}^M\sum_{j=1}^{T} L\big(u_{N, m}(x, t_j; \Theta), u_m(x, t_j)\big),
\label{mean_loss}
\end{equation}
with respect to $\Theta$. Here, $u_m(x, t_j)$ is the solution at $t_j$
of the $m^{\text{th}}$ trajectory in the dataset and $u_{N, m}(x,
t_j;{\Theta})$ denotes the spectral expansion solution reconstructed
from the coefficients $\bm{c}_{N, m}$ obtained by the neural ODE of
the $m^{\text{th}}$ sample at $t_j$.

To solve unbounded domain DEs in $\Omega \subseteq \mathbb{R}^D$, two
additional parameters are needed to scale and translate the spatial
argument $\bm{x}$, a scaling factor $\bm{\beta}\coloneqq
(\beta^1,\ldots,\beta^D)\in \mathbb{R}^D$, and a shift factor
$\bm{x}_0\coloneqq(x_0^1,\ldots,x_0^D)\in\mathbb{R}^D$. These factors,
particularly in unbounded domains, often need to be dynamically
adjusted to obtain accurate spectral approximations of the original
function \citep{tang1993hermite,xia2021efficient,xia2021frequency}.
Note that given the observed $u(X, t)$, the ground truth coefficients
$c_i(t)$ as well as the spectral adjustment parameters $\beta(t)$ and
$x_0(t)$ are obtained by minimizing the \textit{frequency indicator}
(introduced in \citep{xia2021efficient})
\begin{equation}
\mathcal{F}(u; \beta, x_0) = \sqrt{\frac{\sum_{i=N-[\tfrac{N}{3}]+1}^N
  c_i^2}{\sum_{i=0}^N c_i^2}}
\end{equation}
that measures the error in the numerical representation of the
solution $u$ \citep{chou2023adaptive}. Therefore, minimizing
$\mathcal{F}(u; \beta, x_0)$ will also minimize the approximation
error $\|u - \sum_{i=0}^N
c_i\hat{\mathcal{H}}(\beta(t)(x-x_0(t)))\|^2_2$.

Generalizing the spectral approximation Eq.~\eqref{spectral_approx} to
higher spatial dimensions, we can write
\begin{equation}
  u(\bm{x}, t)  \approx u_{N, \bm{x}_0}^{\bm{\beta}}(\bm{x}, t)
=\sum_{i=0}^N c_i(t)\phi_i\big(\bm{\beta} * (\bm{x} - \bm{x}_0)\big),
 \label{spectral_unbounded} 
\end{equation}
where $\bm{\beta}*(\bm{x} - \bm{x}_0) \coloneqq (\beta^1(x-x_0^1),...,
\beta^D(x-x_0^D))$ is the Hadamard product.

The two parameters $\bm{\beta}(t), \bm{x}_0(t)$, which are also
functions of time, can also be learned by the neural ODE. More
specifically, we append the scale and displacement variables to the
coefficient vector $\bm{c}_N(t)$ and write $\tfrac{\dd
  \tilde{\bm{c}}_N}{\dd t} = \bm{F}(\tilde{\bm{c}}_N;t)$ for the ODEs
satisfied by $\tilde{\bm{c}}_N \coloneqq
\big(c_0(t),...,c_N(t),\bm{\beta}(t), \bm{x}_0(t)\big)$.  The
underlying dynamics $\bm{F}(\tilde{\bm{c}}_N;t)$ is approximated as
\begin{equation}
\bm{F}(\tilde{\bm{c}}_N; t)\approx \bm{F}(\tilde{\bm{c}}_N; t, \Theta)
\label{unbounded_dynamics}
\end{equation}
by minimizing with respect to $\Theta$ a loss function that also
penalizes the error in $\beta$ and $x_0$
\begin{equation}
\begin{aligned}
& \sum_{m=1}^M\sum_{j=1}^{T}\bigg[L\big(u_{N, \bm{x}_{0, m}, m}^{\bm{\beta}_m}(\bm{x},
    t_j;{\Theta}),u_m(\bm{x}, t_j)\big) \\[-3pt]
& \hspace{3.7cm} + \lambda \|\bm{\beta}_m(t_j)
  -\bm{\beta}_m(t_j;\Theta)\|_2^2 + \lambda
  \|\bm{x}_{0, m}(t_j)-\bm{x}_{0, m}(t_j;\Theta)\|_2^2\bigg].
\end{aligned}
\label{loss_parameter}
\end{equation}
Similarly, the DE satisfied by $u_{N, \bm{x}_0}^{\bm{\beta}}(\bm{x},
t)$ is
\begin{equation}
\partial_t u_{N, \bm{x}_0}^{\bm{\beta}}(\bm{x}, t)=F[u_{N,
    \bm{x}_0}^{\bm{\beta}};\bm{x},t, \Theta],
\end{equation}
where
\begin{equation}
F[u_{N,\bm{x}_0}^{\bm{\beta}};\bm{x},t,
  \Theta]=\sum_{i=0}^N F_i(\tilde{\bm{c}}_N;t,\Theta)
\phi_i(\bm{\beta}(\bm{x}-\bm{x}_0))
\end{equation}
and $F_i$ is the $i^{\text{th}}$ component of
$\bm{F}(\tilde{\bm{c}}_N; t,\Theta)$.
Here, $\bm{\beta}_{m}(t_j)$ and $\bm{x}_{0, m}(t_j)$ are the scaling
factor and the displacement of the $m^{\text{th}}$ sample at time
$t_j$, respectively, and $\lambda$ is the penalty due to squared
mismatches in the scaling and shift parameters $\beta$ and $x_0$. In
this way, the dynamics of the variables $\bm{x}_0,\bm{\beta}$ are also
learned by the neural ODE so they do not need to be manually adjusted
as they were in
\citep{chou2023adaptive,xia2021efficient,xia2021frequency,xia2022spectrally}.

\added{In case the space $\Omega$ is high-dimensional, and the
  solutions are sufficiently smooth and well-behaved, they can be
  approximated by restricting the basis functions $\{\phi_{i,
    x_0}^{\beta}\}$ to those in the hyperbolic cross space of the full
  tensor product of basis functions. If this projection is performed
  optimally, the effective dimensionality of the problem can be
  reduced \citep{shen2010sparse,shen2010efficient} without significant
  loss of accuracy. We will show that our method can also easily
  incorporate hyperbolic spaces to enhance training efficiency in
  modestly higher-dimensional problems.}

\section{Numerical experiments}
\label{numerical}
Here, we carry out numerical experiments to test our spectral neural
DE method by learning the underlying DE given data in both bounded and
unbounded domains. In this section, all numerical experiments are
performed using the \texttt{odeint\_adjoint} function along with the
\texttt{dopri5} numerical integration method developed in
\citep{chen2018neural} for training the neural network. \added{Both
  stochastic gradient descent (SGD) and the Adam method are used to
  optimize parameters of the neural network.} In this work, we set
$L(\cdot, \cdot)$ to be the relative squared $L^2$ error
\begin{equation}
  L(u(x, t_i),u_N(x, t_i;\Theta))\coloneqq 
  \frac{\|u(x, t_i) - u_N(x, t_i;\Theta)\|_2^2}{\|u\|_2^2}
\label{L2loss}
\end{equation}
in the loss function Eq.~\eqref{mean_loss} used for training.

Since algorithms already exist for learning bounded-domain PDEs, we we
first examine a bounded-domain problem in order to benchmark our
spectral neural DE method against two other recent representative
methods, a convolutional neural PDE learner
\citep{brandstetter2021message} and a Fourier neural operator PDE
learning method \citep{li2020fourier}.



\begin{example}
\rm
\label{example_1}
\added{For our first test example, we consider learning a bounded-domain
  Burgers' equation that might be used to describe the behavior of
  viscous fluid flow \cite{bateman1915some}:}

\begin{equation}
\begin{aligned}
  \partial_t u + \tfrac{1}{2}\partial_x (u^2)
= & \tfrac{1}{10}\partial_{xx} u,\quad x\in(-1, 1),\,\, t\geq 0,\\
  u(-1, t) = u(1, t),\,\, \label{Burgers} 
  \partial_x u(-1, t) & = \partial_x u(1, t), \,\, \,
u(x, 0) = -\frac{1}{5} \frac{\phi_x(x, 0)}{\phi(x, 0)},
\end{aligned}
\end{equation}
where
\begin{equation}
  \phi(x, t) \equiv  5+ \frac{2+\xi_1}{2}e^{-\pi^{2}t/10} \sin \pi x
  + \frac{\xi_2}{2} e^{-2\pi^{2}t/5} \cos 2\pi x.
\label{phi_def}
\end{equation}
This model admits the analytic solution $u(x, t) = -{\phi_x(x,
  t)}{5\phi(x, t)}$.
We then sample the two independent random variables from $\xi_1, \xi_2 \sim
\mathcal{U}(0, 1)$ to generate a class of solutions to
Eq.~\eqref{phi_def} $\{u\}_{\xi_1, \xi_2}$ for both training and
testing. To approximate $F$ in Eq.~\eqref{dynamics}, we use a neural
network that has one intermediate layer with 300 neurons and the ELU
activation function.  The basis functions in
Eq.~\eqref{spectral_approx} are taken to be Chebyshev polynomials.
For training, we use two hundred solutions (each corresponding to a
pair of values $(\xi_{1}, \xi_{2})$ of Eq.~\eqref{Burgers}, record the
expansion coefficients $\{c_i\}_{i=0}^9$ at different times
$t_j=j\Delta{t}, \Delta{t}=\frac{1}{4}, j=0,\ldots,4$.  The test set
consists of another $100$ solutions, also evaluated at times
$t_j=j\Delta{t}, \Delta{t}=\frac{1}{4}, j=0,\ldots,4$.

In this bounded-domain problem, we can compare our results with those
generated from the Fourier neural operator (FNO) and the convolutional
neural PDE learner methods.  In the FNO method, four intermediate
Fourier convolution layers with 128 neurons in each layer were used to
input the initial condition $u(i\Delta x, 0)$ and function values
$u(i\Delta x, t=j\Delta t)$ (with $\Delta x = \frac{1}{128}, \Delta t
= \frac{1}{4}$) \citep{li2020fourier}. In the convolutional neural PDE
learner comparison, we used seven convolutional layers with 40 neurons
in each layer to input $u(i\Delta x,j\Delta t)$ (with
$\Delta{x}=\frac{1}{100}, \Delta{t}=\frac{1}{250}$)
\citep{brandstetter2021message}. Small $\Delta{x}$ and $\Delta{t}$
were used in the convolutional neural PDE learner method because this
method depends on both spatial and temporal discretization, requiring
fine discretization in both dimensions. For all three methods, we used
the Adam method to perform gradient descent with a learning rate
$\eta=0.001$ to run 10000 epochs, which was sufficient for the errors
in all three methods to converge. We list in Table~\ref{tab:burgers}
both the MSE error $\frac{1}{TM}\sum_{m=1}^M\sum_{j=1}^{T}\|u_{N,
  m}(x, t_j; \Theta) - u_m(x,t_j)\|_2^2$ and the mean relative $L^2$
error
\begin{equation}
\frac{1}{TM}\sum_{m=1}^N\sum_{j=1}^T \frac{\|u_{N, m}(x, t_j; \Theta) - u_m(x,
  t_j)\|_2}{\|u_m(x, t_j)\|_2}.
\label{relativeL2}
\end{equation}
\begin{table*}[htb]
  \centering
\small
\renewcommand*{\arraystretch}{1.0}
  \caption{{\small Comparison of different PDE learners. The
      convolutional PDE learner, the Fourier neural operator method,
      and our proposed spectral neural PDE learner are used to learn
      the dynamics of Burgers' equation Eq.~\eqref{Burgers} in a
      \textit{bounded} domain. The FNO method gives the best
      performance and the convolutional neural PDE performs the
      worst. Our proposed spectral neural PDE learner achieves
      performance comparable to the FNO method.}}
  \vspace{5mm}
  \begin{tabular}{l|cc|ccc}
\hline
 &  \multicolumn{2}{c}{Training error}  & \multicolumn{2}{c}{Testing error}  \\
\hline
\diagbox{$H$}{$N_H$} & MSE & Mean relative $L^2$
& MSE & Mean relative $L^2$ 
\\
\hline
 convolutional & 2.39e-05$\pm$1.82e-05 & 1.68e-02$\pm$7.86e-03 &1.10e-04$\pm$6.62e-05 & 2.82e-02$\pm$6.62e-03\\
    Fourier &  3.21e-07$\pm$1.53e-07 &  7.43e-03$\pm$1.76e-03 & 4.66e-07$\pm$3.43e-07 & 8.61e-03$\pm$2.86e-03\\
    spectral  & 1.15e-06$\pm$9.72e-07 & 9.82e-03$\pm$4.95e-03& 1.17e-06$\pm$9.73e-07 &9.99e-03$\pm$4.97e-03\\
    \hline
%
  \end{tabular}%
  \label{tab:burgers}
\end{table*}


For the FNO and spectral PDE learning methods, we
aim to minimize the relative squared loss $L^{2}$
(Eq.~\eqref{L2loss}), while for the convolution neural PDE method, we
must minimize the MSE loss since only partial and local spatial
information on the solution is inputted during each training epoch so
we cannot properly define a relative squared loss.  Thus, as shown in
Table~\ref{tab:burgers}, the MSE of the FNO method is smaller than the
MSEs of the other two methods on the training set while the
convolutional neural PDE method performs the worst. Nonetheless, our
proposed neural spectral PDE learning approach gives similar
performance compared to the FNO method, providing comparable mean
relative $L^2$ and MSE errors for learning the dynamics associated
with the bounded-domain Burgers' equation.

Additionally, the run times in this example were $2$ hours for the
convolutional PDE learning method, 6 hours for the FNO method, and 5
hours for our proposed spectral neural DE learning approach.  These
computations were implemented on a 4-core
Intel\textsuperscript{\textregistered} i7-8550U CPU, 1.80 GHz laptop
using Python 3.8.10, Torch 1.12.1, and Torchdiffeq 0.2.3. Overall,
even in bounded domains, our proposed spectral neural DE learning
approach compares well with the recently developed convolutional
neural PDE and FNO methods, providing comparable
errors and efficiency in learning the dynamics of Eq.~\eqref{Burgers}.


\end{example}

The Fourier neural operator method works well in reconstructing the
Burgers' equation in Example~\ref{example_1}, and there could be other
even more efficient methods for reconstructing bounded domain
spatiotemporal DEs. However, reconstructing unbounded domain
spatiotemporal DEs is substantially different from reconstructing
bounded domain counterparts. First, discretizing space cannot be
directly applied to unbounded domains; second, if we truncate an
unbounded domain into a bounded domain, appropriate artificial
boundary conditions need to be imposed
\cite{antoine2008review}. Constructing such boundary conditions is
usually complex and improper boundary conditions can lead to large
errors, resulting erroneous learning of the DE. A simple example of
when the FNO will fail when we truncate the unbounded domain into a
bounded domain is provided in ~\ref{FNOunbounded}.

Since our spectral method uses basis functions, it obviates the need
for spatial discretization and can be used to reconstruct
unbounded-domain DEs.  Dynamics in unbounded domains are intrinsically
different from their bounded-domain counterparts because functions can
display diffusive and convective behavior leading to, \textit{e.g.},
time-dependent growth at large $x$. This growth poses intrinsic
numerical challenges when using prevailing finite element/finite
difference methods that truncate the domain.
%

Although it is difficult for most existing methods to learn the
dynamics in unbounded spatial domains, our spectral approach can
reconstruct unbounded-domain DEs by simultaneously learning the
expansion coefficients and the evolution of the basis functions.  To
illustrate this, we next consider a one-dimensional unbounded domain
inverse problem.


\begin{example}
\label{example2}
\rm
Consider learning the solution
\begin{equation}
  u(x, t;\xi_1, \xi_2, \xi_3)=  \frac{\xi_1}{\sqrt{t+1+\xi_2}}
\exp\left[-\frac{(x-t - \xi_3)^2}{t+1+\xi_2}\right],\,\,   x\in\mathbb{R},\, t\in[0, 1]
\label{udefinition}
\end{equation}
and its associated parabolic PDE
\begin{equation}
  \partial_t u = -\partial_x u + \frac{1}{4}\partial_{xx} u,\quad
  u(x, 0) = \frac{\xi_1}{\sqrt{1+\xi_2}}\exp\left[-\frac{(x-\xi_3)^2}{\xi_2+1}\right].
\label{conv_diffu}
\end{equation}
This problem is defined on an unbounded domain and thus neither the
FNO nor the convolutional neural PDE methods can
be used as they rely on spatial discretization (and thus bounded
domains). However, given observational data $u(\cdot, t)$ for
different $t$, we can calculate the spectral expansion of $u$ via the
generalized Hermite functions \citep{Spectral2011}
\begin{equation}
u(x, t)\approx u_{N, x_0}^{\beta}=\sum_{i=0}^N c_i(t) 
\hat{\mathcal{H}}_i\big(\beta(t)(x-x_0(t))\big)
\label{spec_expan}
\end{equation}
and then use the spectral neural DE learning approach to reconstruct
the dynamics $F$ in Eq.~\eqref{objective} satisfied by $u$.  Recall
that the scaling factor $\beta(t)$ and the displacement of the basis
functions $x_0(t)$ are also to be learned.  To penalize misalignment
of the spectral expansion coefficients and the scaling and
displacement factors $\beta$ and $x_0$, we use the loss function
Eq.~\eqref{loss_parameter}.  Note that taking the derivative of
$\beta(t_j; \Theta)$ with respect to $\Theta$ in the first term of
Eq.~\eqref{loss_parameter} would involve evaluating ~$\int
\partial_{\Theta}\beta(t_j;\Theta)(x-x_0(t_j;\Theta))u_{N,
  x_0}^{\beta}(t_j; \Theta) \cdot\partial_x u_{N, x_0}^{\beta}(x, t_j;
\Theta) \text{d}{x}$; similarly, taking the derivative of
$x_0(t_j;\Theta)$ with respect to $\Theta$ would involve evaluating
$\int \partial_{\Theta}x_0(t_j;\Theta)\beta(t_j;\Theta)u_{N,
  x_0}^{\beta}(t_j; \Theta) \partial_x u_{N, x_0}^{\beta}(x, t_j;
\Theta)\text{d}{x}$.
Expressing $\partial_x u_{N, x_0}^{\beta}(x, t_j; \Theta)$ in terms of
the basis functions $\phi_{i, x_0}^{\beta}$ would involve a dense
matrix-vector multiplication of the coefficients of the expansion
$\partial_x u_{N, x_0}^{\beta}(x, t_j; \Theta)$, which might be
computationally expensive during backward propagation in the
stochastic gradient descent (SGD) procedure. Alternatively,
\added{suppose the parameter set of the neural network is
  $\Theta_{j-1}$ before the $j^{\text{th}}$ training epoch. We can fix
  $\tilde{\beta}(t_j)\coloneqq \beta(t_j;\Theta_{j-1}),
  \tilde{x}_0(t_j) \coloneqq x_0(t_j;\Theta_{j-1})$ ($\tilde{\beta},
  \tilde{x}_0$ are fixed values and do not depend on the weights
  $\Theta$) and then modify Eq.~\eqref{loss_parameter} to}
\begin{equation}
\begin{aligned}
&\sum_{m=1}^M \sum_{j=1}^{T} \bigg[
\frac{\|u_{N, \tilde{x}_{0, m}(t_j)}^{\tilde{\beta}_m(t_j)}(x, t_j;\Theta) 
- u_m(x, t_j)\|_2^2}{\|u_m(x, t_j)\|_2^2} \\[-2pt]
& \hspace{3.6cm} +\lambda\big(\beta_m(t_j; \Theta)
- \beta_m(t_j)\big)^2 + \lambda  \big(x_{0, m}(t_j; \Theta) 
- x_{0, m}(t_j)\big)^2\bigg], 
\end{aligned}
\label{L2unboundedloss_new}
\end{equation}
so that backpropagation within each epoch will \textit{not} involve
taking derivatives w.r.t. $\Theta$ for $\tilde{\beta}_m(t_j),
\tilde{x}_{0, m}(t_j)$ in the first term of
Eq.~\eqref{L2unboundedloss_new}.  \added{The reason is that when
  $\beta_m(t), x_{m, 0}(t)$ are well fitted (\textit{e.g.}
  $\beta_m(t;\Theta_{j-1})=\beta(t), x_{m, 0}(t;\Theta_{j-1})=x_0(t)$
  are ground-truth values), Eq.~\eqref{L2unboundedloss_new} will
  simply become}

\begin{equation}
\begin{aligned}
\sum_{m=1}^M \sum_{j=1}^{T}
\frac{\|u_{N, \tilde{x}_{0, m}(t_j)}^{\tilde{\beta}_m(t_j)}(x, t_j;\Theta) 
- u_m(x, t_j)\|_2^2}{\|u_m(x, t_j)\|_2^2}
=\sum_{m=1}^M \sum_{j=1}^{T} \frac{\sum_{i=0}^N (c_{m, i}(t_j;\Theta)
-c_i(t_j))^2}{\sum_{i=0}^N c_{m, i}(t_j)^2}
\end{aligned}
\end{equation}
\added{so no derivative of $\beta, x_0$ w.r.t. $\Theta$ will be used
  and only the gradient of $F$ in Eq.~\eqref{dynamics} w.r.t. $\Theta$
  arises.  Therefore, we can consider separating the fitting of the
  coefficients $c_i(t)$ and the fitting of $\beta(t), x_0(t)$.}


We use 100 samples for training and another 50 samples for testing
with $N=9, \Delta{t}=0.1, T=9, \lambda=0.1$. A neural network with two
hidden layers, 200 neurons in each layer, and the ELU activation
function is used for training. Both training and testing data are
taken from Eq.~\eqref{udefinition} with sampled parameters
$\xi_1\sim\mathcal{N}(3,\tfrac{1}{4}),\,\, \xi_2\sim \mathcal{U}(0,
\tfrac{1}{2}),\,\, \xi_3\sim\mathcal{N}(0, \tfrac{1}{2})$.

Setting $\lambda=0.1$, we first compare the two different loss
functions Eqs.~\eqref{loss_parameter} and~\eqref{L2unboundedloss_new}.
After running 10 independent training processes using SGD, each
containing 2000 epochs and using learning rate $\eta=0.0002$, the
average relative $L^2$ error when using the loss function
Eq.~\eqref{loss_parameter} are larger than the average relative $L^2$
errors when using the loss function Eq.~\eqref{L2unboundedloss_new}.
This difference arises in both the training and testing sets as shown
in Fig.~\ref{example2_fig}(a).
%
\begin{figure}[htb]
\centering
\includegraphics[width=0.75\textwidth]{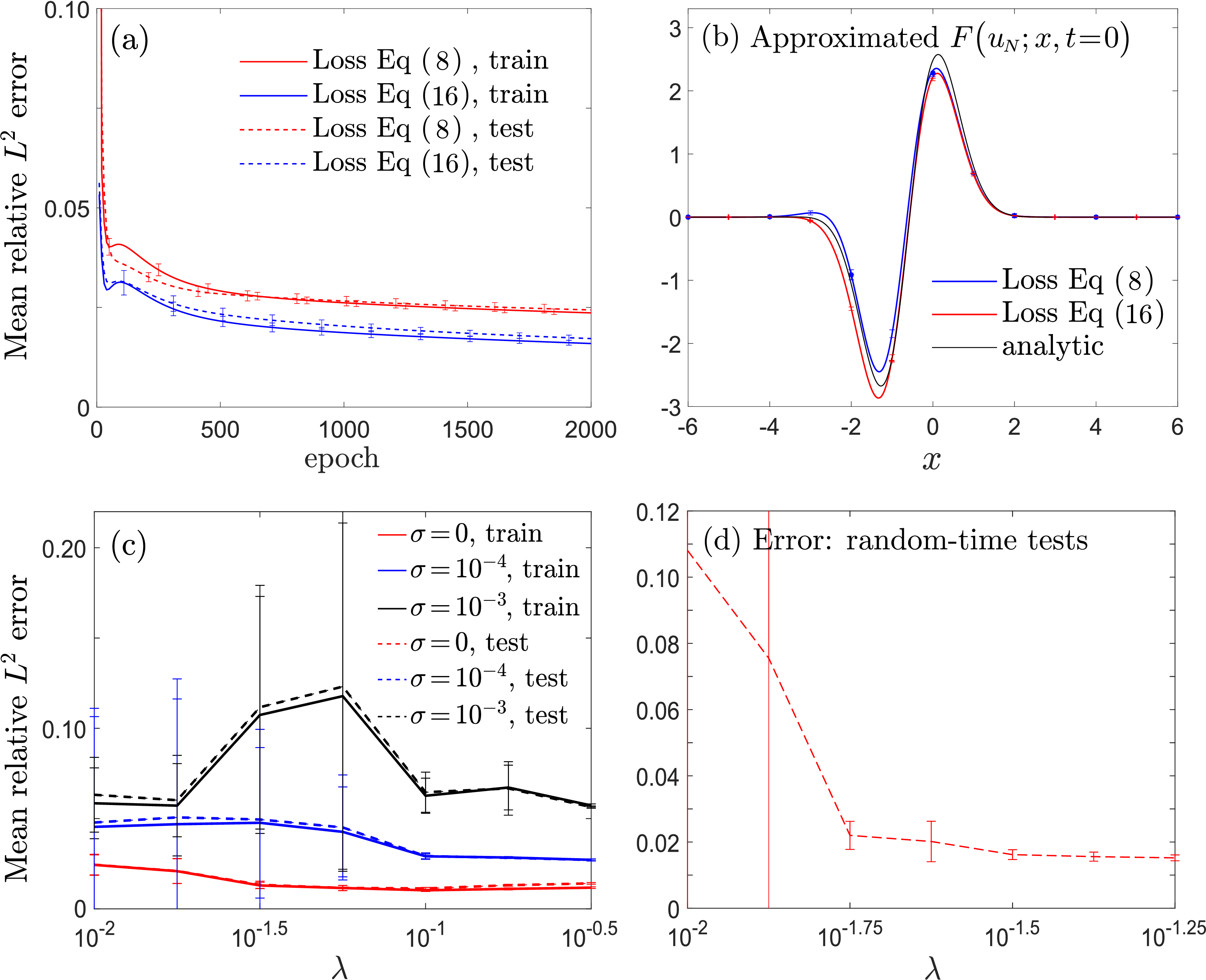}
\caption{\small (a) Errors using different loss functions
  Eqs.~\eqref{loss_parameter} and~\eqref{L2unboundedloss_new}. (b)
  Average dynamics found from using Eqs.~\eqref{loss_parameter}
  and~\eqref{L2unboundedloss_new}.  (c) Errors with $\lambda$ and
  $\sigma$. (d) Errors on the testing set with random times
  $t_i\sim\mathcal{U}(0, 1.5)$.}
\label{example2_fig}
\vspace{-0.15in}
\end{figure}
%

In Fig.~\ref{example2_fig}(b), we plot the average learned $F$ (RHS in
Eq.~\eqref{objective}) for a randomly select sample at $t=0$ in the
testing set. The dynamics learned by using
Eq.~\eqref{L2unboundedloss_new} is a little more accurate than that
learned by using Eq.~\eqref{loss_parameter}.  Also, using the loss
function Eq.~\eqref{L2unboundedloss_new} required only $~\sim 1$ hour
of computational time compared to 5 days when learning with
Eq.~\eqref{loss_parameter} (on the 4-core i7-8550U laptop).
%
%
Therefore, for efficiency and accuracy, we adopt the revised loss
function Eq.~\eqref{L2unboundedloss_new} and separately fit the
dynamics of the adaptive spectral parameters ($\beta, x_0$) and the
spectral coefficients $c_{i}$.

We also explore how network architecture and regularization affect the
reconstructed dynamics.  The results are shown in ~\ref{structured},
from which we observe that a wider and shallower neural network with 2
or 4 intermediate layers and 200 neurons in each layer yields the
smallest errors on both the training and testing sets, \textit{and}
short run times.  We also apply a ResNet \citep{He2016jun} as well as
dropout \citep{hinton2012improving, Srivastava2014DropoutAS} to
regularize the neural network structure. Dropout regularization does
not reduce either the training error or the testing error probably
because even with a feedforward neural network, the errors from our
spectral neural DE learner on the training set are close to those on
the testing set and there is no overfitting issue.  On the other hand,
applying the ResNet technique leads to about a ~20\% decrease in
errors.
%
%
Results from using ResNets and dropout are shown in ~\ref{structured}.

Next, we investigate how noise in the observed data and changes in the
adaptive parameter penalty coefficient $\lambda$ in
Eq.~\eqref{L2unboundedloss_new} impact the results.  Noise is
incorporated into simulated observational data as
\begin{equation}
u_{\xi}(x, t) = u(x, t) [1+ \xi(x, t)],
\label{conv_diff}
\end{equation}
where $u(x, t)$ is the solution to the parabolic equation
Eq.~\eqref{conv_diffu} given by Eq.~\eqref{udefinition} and $\xi(x,
t)\sim\mathcal{N}(0, \sigma^2)$ is a Gaussian-distributed noise that
is both spatially and temporally uncorrelated (\textit{i.e.}, $\langle
\xi(x, t)\xi(t, s)\rangle=\sigma^2\delta_{x, y}\delta_{s, t}$). The
noise term is assumed to be independent for different samples.  We use
a neural network with 2 hidden layers, 200 neurons in each layer, to
implement 10 independent training processes using SGD and a learning
rate $\eta=0.0002$, each containing 5000 epochs. Results are shown in
Fig.~\ref{example2_fig}(c) and further tabulated in~\ref{noise_result}.
For $\sigma=0$, choosing an intermediate $\lambda\in (10^{-1.5},
10^{-1}]$ leads the smallest errors and an optimal balance between
learning the coefficients $c_{i}$ and learning the dynamics of $\beta,
x_0$.  When $\sigma$ is increased to nonzero values ($\sim 10^{-4} -
10^{-3}$), a larger $\lambda \sim 10^{-0.75}-10^{-0.5}$ is needed to
keep errors small (see Fig.~\ref{example2_fig}(c)
and~\ref{noise_result}).  If the noise is further increased to, say,
$\sigma=10^{-2}$ (not shown in Fig.~\ref{example2_fig}), an even
larger $\lambda \sim 10^{-0.5}$ is needed for training to converge.
This behavior arises because the independent noise $\xi(x,
t)\sim\mathcal{N}(0, \sigma^2)$ contributes more to high-frequency
components in the spectral expansion. In order for training to
converge, fitting the shape of the basis functions by learning $\beta,
x_0$ is more important than fitting noisy high-frequency components
via learning $c_{i}$.  A larger $\lambda$ puts more weight on learning
the dynamics of $\beta, x_0$ and basis function shapes.

We also investigate how the intrinsic noise in the parameters $\xi_1,
\xi_2, \xi_3$ of the solution's distribution Eq.~\eqref{udefinition}
affect the accuracy of the learned DE by our method. \added{We found
  that if the intrinsic noise in $\xi_1, \xi_2, \xi_3$ is larger, then
  the training errors of the learned DE models are larger. However,
  compared to models trained on data with lower $\xi_1, \xi_2, \xi_3$,
  training using noisier data leads to lower errors when testing data
  are also noisy. Training with noisier data leads to better
  performance when testing data are also noisy.  The training and
  testing errors that show this feature are presented in
  \ref{intrinsicnoise}.}




Finally, we test whether the parameterized $F$
(Eq.~\eqref{unbounded_dynamics}) learned from the training set can
extrapolate well beyond the training set sampling interval $t\in[0,
  0.9]$. To do this, we generate another 50 trajectories and sample
each of then at random times $t_i\sim\mathcal{U}(0, 1.5),
i=1,...,9$. We then used models trained with $\sigma=0$ and different
$\lambda$ to test.
%
%
As shown in Fig.~\ref{example2_fig}(d), our spectral neural DE learner
can accurately extrapolate the solution to times beyond the training
set sampling time intervals. We also observe that a stronger penalty
on $\beta$ and $x_0$ ($\lambda=10^{-0.5}$) leads to better
extrapolation results.

\end{example}

In the last example, we carry out a numerical experiment on learning
the evolution of a Gaussian wave packet (which may depend on nonlocal
interactions) across a two-dimensional unbounded domain $(x,
k)\in\mathbb{R}^2$. We use this case to explore improving training
efficiency by using a hyperbolic cross space to reduce the number of
coefficients in multidimensional problems.

\begin{example}
\label{wigner_example}
\rm We solve a 2D unbounded-domain problem
of fitting a Gaussian wave packet's evolution
\begin{equation}
  f(x, k, t; \xi_1, \xi_2) = 2e^{-\frac{(x-\xi_1)^2}{2a^2}}\,
  e^{2\beta t (x-\xi_{1})(k-\xi_2)}\, e^{-2a^2(1+\beta^2t^2)(k-\xi_2)^2},
\label{wigner_solution}
\end{equation}
where $\xi_1$ is the center of the wave packet and $a$ is the minimum
positional spread. \added{If $\xi_2=0$, the Gaussian wave packets
  defined in Eq.~\eqref{wigner_solution} solves the stationary
  zero-potential Wigner equation, an equation often used in quantum
  mechanics to describe the evolution of the Wigner quasi-distribution
  function \citep{chen2019numerical,shao2011adaptive}.}
We set $a=1$ and $\beta=1/2$ in Eq.~\eqref{wigner_solution} and
independently sample $\xi_1, \xi_2\sim\mathcal{U}(-1/2, 1/2)$ to
generate data. Thus, the DE satisfied by the highly nonlinear
Eq.~\eqref{wigner_solution} is unknown and potentially involves
nonlocal convolution terms. In fact, there could be infinitely many
DEs, including complicated nonlocal DEs, that can describe the
dynamics of Eq.~\eqref{wigner_solution}. An example of such a nonlocal
DE is

\begin{equation}
\begin{aligned}
\partial_t f + & 2a^2\beta A[f; x,k, t]\partial_{x} f(x, k, t)  = 0, \\[4pt]
A[f; x,k, t] & =  \frac{\beta t (x - B[f;x,k,t])}{2a^2(1+\beta^2 t^2)}  
+ \sqrt{\tfrac{\log D[f; k, t] - C(t) - \log(f/2)}{a^2(1+\beta^2t^2)}},\\[3pt]
B[f; x, k, t] & = x - \sqrt{2a^2(1+\beta^2t^2)}\times  \sqrt{2C(t)
  - 2\log D[f; x, k, t] +\log (f/2)},\\[3pt]
C(t) & = \tfrac{1}{2}\log\left[\frac{\pi}{a^{2}(1+\beta^{2}t^{2})}\right],\\
D[f;x, k, t] & = \int f(x, y, t)e^{-2a^2(1+\beta^2t^2)(y-k)^2} \text{d}y.
\end{aligned}
\end{equation}
We wish to learn the underlying dynamics using a parameterized $F$ in
Eq.~\eqref{unbounded_dynamics}.  Since the Gaussian wave packet
Eq.~\eqref{wigner_solution} is defined in the unbounded domain
$\mathbb{R}^2$, learning its evolution requires information over the
entire domain. Thus, methods that depend on discretization of space
are not applicable.

Our numerical experiment uses Eq.~\eqref{wigner_solution} as both
training and testing data. We take $\Delta{t}=0.1, t_j=j\Delta{t},
j=0,...,10$ and generate 100 trajectories for training. For this
example, training with ResNet results in diverging gradients, whereas
the use of a feedforward neural network yields convergent results. So
we use a feedforward neural network with two hidden layers and 200
neurons in each hidden layer and the ELU activation function. We train
across 1000 epochs using SGD with momentum (SGDM), a learning rate
$\eta=0.001$, $\text{momentum}=0.9$, and $\text{weight decay}=0.005$.
For testing, we generate another 50 trajectories, each with starting
time $t_0=0$ but $t_j$ taken from $\mathcal{U}(0, 1), j=1,\dots,10$.
We use the spectral expansion
\begin{equation}
f_N(x, k, t_j; \xi_1, \xi_2)= \sum_{i=0}^{14}\sum_{\ell=0}^{14}
    c_{i, \ell}(t_j)\hat{\mathcal{H}}_i(\beta^1(x-x_0^1))
    \hat{\mathcal{H}}_{\ell}(\beta^2(k-k_0^2))
\label{f_expand}
\end{equation}
to approximate Eq.~\eqref{wigner_solution}.  We record the
coefficients $c_{i, \ell}$ as well as the scaling factors and
displacements $\beta^{1}, \beta^{2}, x_0^1, k_0^2$ at different $t_j$.

Because $(x, k)\in \mathbb{R}^2$ are defined in a 2-dimensional space,
instead of a tensor product, we can use a hyperbolic cross space for
the spectral expansion to effectively reduce the total number of basis
functions while preserving accuracy \citep{shen2010sparse}. Similar to
the use of sparse grids in the finite element method
\citep{bungartz2004sparse,zenger1991sparse}, choosing basis functions
in the space
\begin{equation}
\begin{aligned}
  & V_{N, \gamma}^{\vec{\beta}, \vec{x}_0} \coloneqq \text{span}
  \Big\{\hat{\mathcal{H}}_{n_1}(\beta^1(x-x_0^1))
  \hat{\mathcal{H}}_{n_2}(\beta^2(k-k_0^2)): |\vec{n}|_{\text{mix}}\|
  \vec{n}\|_{\infty}^{-\gamma}\leq N^{1-\gamma} \Big\},\\ &
  \vec{n}\coloneqq(n_1, n_2), ~|\vec{n}|_{\text{mix}} \coloneqq
  \max\{n_1, 1\}\cdot \max\{n_2, 1\}
\end{aligned}
\label{hyper_space} 
\end{equation}
can reduce the effective dimensionality of the problem.  We explored
different hyperbolic spaces $V_{N, \gamma}^{\vec{\beta}, \vec{x}_0}$
with different $N$ and $\gamma$. We use the loss function
Eq.~\eqref{L2unboundedloss_new} with $\lambda=\tfrac{1}{50}$ for
training. The results are listed in~\ref{appendix_gamma}.
%
%
To show how the loss function Eq.~\eqref{L2unboundedloss_new} depends
on the coefficients $c_{i, \ell}$ in Eq.~\eqref{f_expand}, we plot
saliency maps \citep{Simonyan2013DeepIC} for the quantity
$\tfrac{1}{10}\sum_{j=1}^{10}
\big|\frac{\partial\text{Loss}_j}{\partial c_{i,
    \ell}(0)}\big|$\footnote{We take derivatives w.r.t. to
  only the coefficients $\{c_{i, \ell}(0)\}$ of $u_{N, \tilde{x}_0(0),
    m}^{\tilde{\beta}_m(0)}(x, 0;\Theta)$ in
  Eq.~\eqref{L2unboundedloss_new} and not with w.r.t. the expansion
  coefficients of the observational data $u(x, 0)$.}, the absolute
value of the partial derivative of the loss function
Eq.~\eqref{L2unboundedloss_new} w.r.t. $c_{i, \ell}$ averaged over 10
training processes.
\begin{figure}[htb]
\begin{center}
\includegraphics[width=0.99\textwidth]{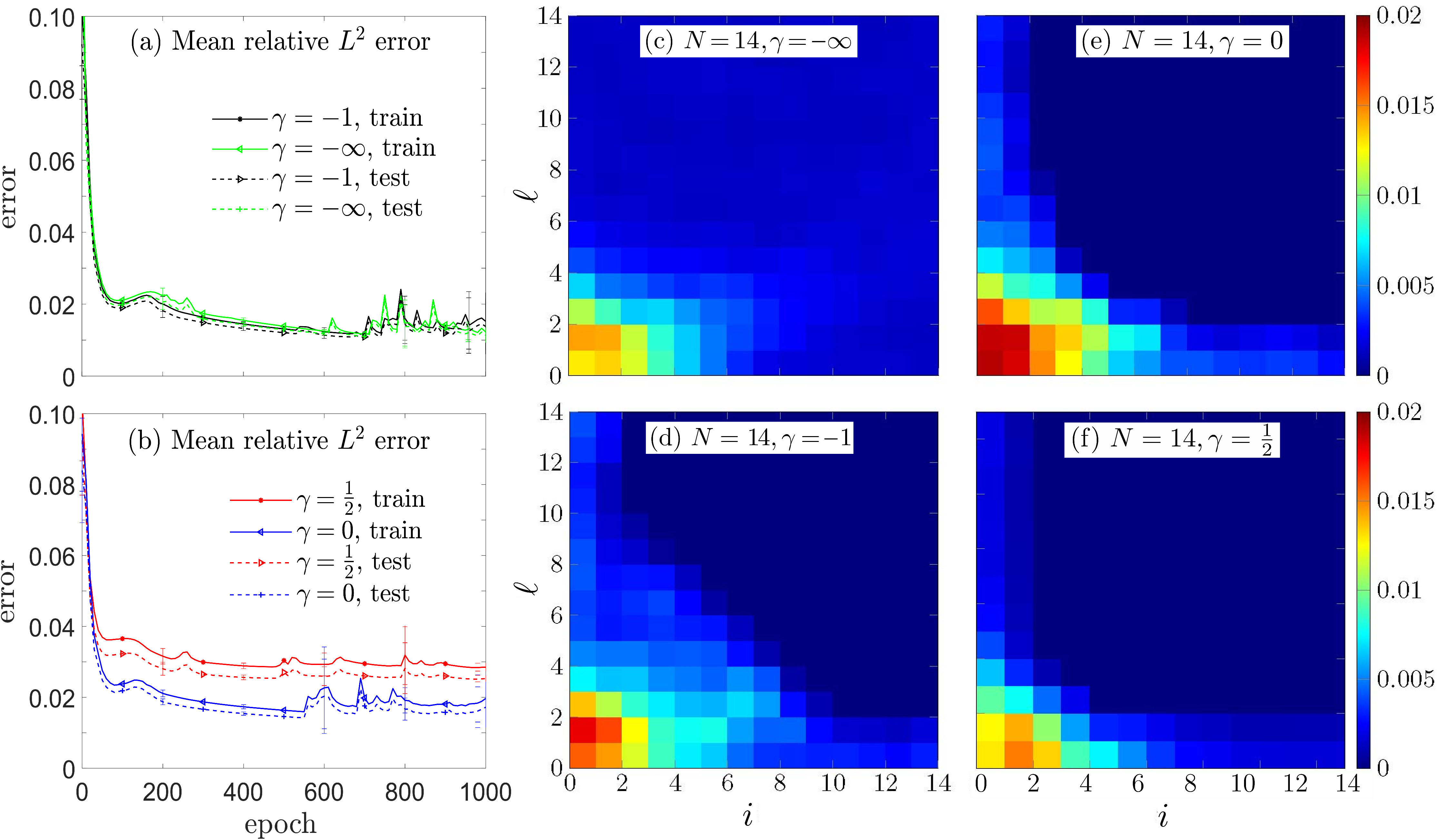}
\vspace{-0.05in}
\caption{\small (a,b) Mean relative $L^2$ errors for $N=14$ and
  $\gamma=-\infty, -1, 0, 1/2$.  (c-f) Saliency maps showing the mean
  absolute values of the partial derivative of the loss function
  w.r.t. to $\{c_{i, \ell}(0)\}$ for $\gamma =-\infty, -1,0,1/2$.}
\label{wigner_map}
\end{center}
\vspace{-0.2in}
\end{figure}
As shown in Fig.~\ref{wigner_map}(a,b), using $\gamma=-1, 0$ leads to
similar errors as the full tensor product $\gamma=-\infty$, but could
greatly reduce the number of coefficients and improve training
efficiency. Taking a too large $\gamma=1/2$ leads to larger errors
because useful coefficients are left out. From
Fig.~\ref{wigner_map}(c-f), there is a resolution-invariance for the
dependence of the loss function on the coefficients $c_{i, \ell}$
though using different hyperbolic spaces with different $\gamma$. 
%
%
We find that an intermediate $\gamma\in(-\infty, 1)$ (\textit{e.g.},
$\gamma=-1, 0$) can be used to maintain accuracy and reduce the number
of inputs/outputs when reconstructing the dynamics of
Eq.~\eqref{wigner_solution}. Overall, the ``curse of dimensionality"
can be mitigated by adopting a hyperbolic space for the spectral
representation.
\end{example}
Finally, in~\ref{rec_appendix}, we consider source reconstruction in a
heat equation. Our proposed spectral neural DE learning method
achieves an average relative error $L^2 \approx 0.1$.  On the other
hand, if all terms on the RHS of Eq.~\eqref{objective} except an
unknown source (which does not depend on the solution) is known, the
recently developed s-PINN method \citep{xia2022spectrally} achieves a
higher accuracy.  However, if in addition to the source term,
additional terms on on the RHS of Eq.~\eqref{objective} are unknown,
s-PINNs cannot be used but our proposed spectral neural DE learning
method remains applicable.

\section{Conclusion}
\label{summary}
In this paper, we proposed a novel spectral neural DE learning method
that is suitable for learning spatiotemporal DEs from spectral
expansion data of the underlying solution.  Its main advantages is
that it can be applied to learning both spatiotemporal PDEs and
integro-differential equations in \textit{unbounded domains}, while
matching the performance of the most recent high-accuracy PDE learning
methods applicable to only bounded domains. Moreover, our proposed
method has the potential to deal with higher-dimensional problems if a
proper hyperbolic cross space can be justified to effectively reduce
the dimensionality.

As a future direction, we plan to apply our spectral neural DE
learning method to many other inverse-type problems in physics.  One
possible application is to learn the stochastic dynamics associated
with anomalous transport \cite{jin2015tutorial} in an unbounded
domain, which could be associated with fractional derivative or
integral operators in the corresponding $F[u; x, t]$ term.  Finally,
since the number of inputs (expansion coefficients) grows
exponentially with spatial dimension, very high dimensional problems
may not be sufficiently mitigated by even the most optimal cross space
hyperbolicity $\gamma$ (see Eq.~\eqref{hyper_space}). Thus, higher
dimensional problems remain challenging and exploring how prior
knowledge on the observed data can be used within our spectral
framework to address them may be a fruitful avenue of future
investigation.

\section*{Acknowledgments}
This work was supported in part by the Army Research Office under
Grant W911NF-18-1-0345.

\bibliography{bibliography}
\bibliographystyle{elsarticle-num}

\newpage

\appendix

\section{Using Fourier neural operator to reconstruct unbounded domain DEs}
\label{FNOunbounded}
We shall show through a simple example that it is usually difficult to
apply bounded domain DE reconstructing methods to reconstruct
unbounded domain DEs even if we truncate the unbounded domain into a
bounded domain, because appropriate boundary conditions must be
provided.  We wish to use the Fourier neural operator method to
reconstruct the unbounded domain DE

\begin{equation}
u_t = \frac{1}{4}u_{xx},\quad  x\in \mathbb{R},\,\,  t\in[0, 1].
\label{heat_append}
\end{equation}
If one imposes the initial condition $u(x, 0) = 10\xi \exp(-100x^2)$,
then
\begin{equation}
u(x,t) = \frac{\xi}{\sqrt{0.01+t}}\exp(-\tfrac{x^2}{0.01+t})
\end{equation}
is the analytic solution to Eq.~\eqref{heat_append}.  For this
problem, we will assume $\xi\sim\mathcal{U}(1, \frac{3}{2})$.

Since the Fourier neural operator (FNO) method relies on spatial
discretization and grids, and cannot be directly applied to unbounded
domain problems, we truncate the unbounded domain.  Suppose one is
interested in the solution's behavior for $x\in[-1, 1]$.  One approach
is to truncate the unbounded domain $x\in\mathbb{R}$ to $[-1, 1]$ and
use the FNO method to reconstruct the solution $u(x, t),\, x\in[-1,
  1], t\in[0, 1]$ given $u(x, 0)$. However, we show how improper
boundary conditions of the truncated domain can leads to large errors.

For example, we assume the training set satisfies the boundary
condition $u(x=\pm 1, t) = 0$, which is not the correct boundary
condition \added{since it is inconsistent with the ground truth
  solution.}  Therefore, we will \textbf{not} be reconstructing the
model in Eq.~\eqref{heat_append}.  We generated the testing dataset
using the correct initial condition $u(x, 0) = 10\xi \exp(-100x^2)$,
without boundary conditions. The results are given in
Table~\ref{tab:bc}.
\begin{table*}[h]
\caption{{\small Training and testing errors when using the FNO
    method and truncating the unbounded domain. However, an incorrect
    boundary condition for the training set is imposed. The error is
    significantly larger on the testing set than that on the training
    set because a different DE is reconstructed from the training
    data.}}  \small
\vspace{5mm}
    \begin{tabular}{lcc|ccc}
\toprule
 &  \multicolumn{2}{c}{Training error}  & \multicolumn{2}{c}{Testing error}  \\[2pt]
\hline
 & \!\!MSE\!\! & \!\!Mean relative $L^2$\!\! 
& \!\!MSE & Mean relative $L^2$ \!\! \\[2pt]
\hline
    Fourier &  3.78e-05 $\pm$ 1.78e-05 &  6.09e-03$\pm$1.47e-03 
    & 8.95e-04$\pm$6.95e-05 & 3.07e-02$\pm$1.17e-03\\[2pt]
    \hline
    \end{tabular}%
  \centering
  \label{tab:bc}
\end{table*}
From Table~\ref{tab:bc}, the testing error is significantly larger
than the training error because a different DE (not
Eq.~\eqref{heat_append}) is constructed from the training data, which
is not the heat equation we expect. Therefore, even if methods such
FNO are efficient in bounded domain DE reconstruction problems,
directly using them to reconstruct unbounded domain problems is not
feasible if we cannot construct appropriate boundary conditions.

\section{Dependence on neural network architecture}
\label{structured}
The neural network structure of the parameterized $F[u;x,t,\Theta]$
may impact learned dynamics. To investigate how the neural network
structure influences results, we use neural networks with various
configurations to learn the dynamics of Eq.~\eqref{udefinition} in
Example~\ref{example2} in the noise-free limit.  We set the learning
rate $\eta=0.0002$ and apply networks with 2,3,5,8 intermediate
layers, and 50, 80, 120, 200 neurons in each layer.

\begin{table}[h]
    \caption{{\small The relative $L^2$ errors on the training set and
        testing set (in parentheses) for Example 2 when there is no
        noise in both the training data and the testing data
        ($\sigma=0$). $\lambda=0.1$ in
        Eq.~\eqref{L2unboundedloss_new}, and the training rate is set
        to be $\eta=0.0005$ for 5000 training epochs using SGD.}}
    \small
\vspace{4mm}
\renewcommand*{\arraystretch}{1.2}
    \begin{tabular}{l|cccc}
      \hline
\diagbox{Neurons}{Layers} & 2 & 3 & 5 & 8\\[2pt]
\hline
    50 &  $0.0166$ $(0.0179)$  & {$0.0175$} {$(0.0196)$} &{$0.0219$} {$(0.0251)$} &{$0.0249$} {$(0.0272)$} \\[2pt]
    80 &   {$0.0143$} {$(0.0155)$}&{$0.0164$} {$(0.0181)$} & {$0.0186$ } {$(0.0208)$} & {$0.0238$} {$(0.0266)$}\\[2pt]
120  &   {$0.0130$} {$(0.0141)$}&{$0.0148$} {$(0.0162)$} &{$0.0192$} {$ (0.0218)$} &{$0.0232$} {$(0.0265)$}\\[2pt]
200  &   {$0.0098$} {$(0.0108)$}& {$0.0126$} {$(0.0137)$} &{$0.0176$} {$ (0.0196)$} &  {$0.0228$} {$(0.0263)$}\\[2pt]
\hline
    \end{tabular}%
    \centering
\label{tab:parabolic}
\end{table}

\begin{table}[htb]
    \caption{{\small The training time (in seconds) for Example 2 when
        there is no noise in both the training data and the testing
        data ($\sigma=0$). $\lambda=0.1$ in
        Eq.~\eqref{L2unboundedloss_new}, and the training rate is set
        to be $\eta=0.0005$ for 5000 training epochs using
        SGD. Training was performed on a laptop with a 4-core
        Intel\textsuperscript{\textregistered} i7-8550U CPU @ 1.80 GHz
        using Python 3.8.10, Torch 1.12.1, and Torchdiffeq 0.2.3.}}
    \small
\vspace{4mm}
\renewcommand*{\arraystretch}{1.2}
    \begin{tabular}{l|cccc}
\hline
\diagbox{Neurons}{Layers} & 2 & 3 & 5 & 8\\[2pt]
\hline
50 &  {$5306\pm216$}  & {$5447\pm885$} & {$5780\pm1027$} & {$6110\pm1233$}\\[2pt]
80 &   {$5204\pm468$}&{$5415\pm311$} & {$6291\pm923$} &{$5717\pm530$}\\[2pt]
120  &  {$5860\pm491$}&{$6286\pm444$} &{$6114\pm872$} & {$7098\pm1141$}\\[2pt]
200  &   {$5438\pm522$}& {$6282\pm672$} & {$6640\pm741$} & {$9282\pm217$}\\[2pt]
\hline
    \end{tabular}%
    \centering
\label{tab:parabolic_time}
\end{table}

From Tables~\ref{tab:parabolic} and~\ref{tab:parabolic_time}, we see
that a shallower and wider neural network yields the smallest
error. Runtimes increase with the number of layers and the number of
neurons in each layer; however, when the number of layers is small,
the increase in runtime with the number of neurons in each layer is
not significant. Thus, for the best accuracy and computational
efficiency, we recommend a neural network with 2 hidden layers and 200
neurons in each layer.

Regularization of the neural network can also affect the spectral
neural PDE learner's ability to learn the dynamics or to reduce
overfitting on the training set.  We set $\lambda=0.1$ in the loss
function Eq.~\eqref{L2unboundedloss_new} and the training rate
$\eta=0.0002$ and train over 5000 epochs using SGD. We applied the
ResNet and dropout techniques with a neural network containing 2
intermediate layers, each with 200 neurons.  For the ResNet technique,
we add the output of the first hidden layer to the output of the
second hidden layer as the new output of the second hidden layer. For
the dropout technique, each neuron in the second hidden layer is set
to 0 with a probability $p$. The results are presented in
Table~\ref{tab:resnet} below.

\begin{table*}[htb]
    \caption{\small Mean relative $L^2$ errors and their standard
      deviations for the training and testing (in parentheses) sets
      for a noiseless ($\sigma=0$) Example~\ref{example2}.  The ResNet
      and dropout techniques are applied to regularization the neural
      network used to parameterize $F[u; x, t]\approx F[u;x, t,
        \Theta]$. For dropout, we experimented with different
      probabilities $p$ of dropping out the links between neurons. The
      errors are averaged over 10 independent training processes.}
    \small
    \vspace{4mm}
    \renewcommand*{\arraystretch}{1.2}
    \begin{tabular}{l|cccc}
\hline
Method & None & ResNet   \\[2pt]
\hline
     & {$0.0101\pm0.0006$} {$(0.0112\pm0.0005)$}&{$0.0083\pm0.0005$} {$ (0.0099\pm0.0004)$} \\[2pt]
\hline
\diagbox{$p$}{Method} &  Dropout & Dropout \& ResNet \\[2pt]
\hline
   0.1   & {$0.0146\pm0.0006$} {$ (0.0153\pm0.0007)$}&{$0.0125\pm0.0004$} {$ (0.0139\pm0.0004)$}\\[2pt]
   0.5  & {$0.0314\pm0.0021$} {$ (0.0327\pm0.0023)$}& {$0.0275\pm0.0018$} {$ (0.0286\pm0.0021)$}\\[2pt]
\hline
    \end{tabular}%
\centering
    \label{tab:resnet}
\end{table*}

Table~\ref{tab:resnet} shows the relative $L^2$ errors on the training
set and testing set for Example 2 when there is no noise in both the
training and testing data ($\sigma=0$). We apply regularization to the
neural network, testing both the ResNet and the dropout techniques
with different dropout probabilities $p$. The errors are averaged over
10 independent training processes. Applying the ResNet technique leads
to approximately a 20\% decrease in the errors, whereas applying the
dropout technique does not reduce the training error nor the testing
error.
\section{How data noise and penalty parameter $\lambda$ affect learning}
\label{noise_result}
We now investigate how different strengths $\sigma$ and penalties
$\lambda$ affect the learning, including the dynamics of $\beta$ and
$x_0$.  For each strength of noise $\sigma=0, 0.0001, 0.001$, 100
trajectories are generated for training and another 50 are generated
for testing according to Eq.~\eqref{conv_diff}. The penalty parameter
tested are $\lambda = 10^{-2}, 10^{-1.5}, 10^{-1}, 10^{-0.5}$.  The
mean relative errors on the training set and testing set over 10
independent training processes are shown in Table~\ref{tab:noise}
below and are plotted in Fig.~\ref{example2}(c).
 
\begin{table*}[h]
\vskip -0.12in
  \caption{{\small Mean relative $L^2$ errors defined in
      Eq.~\eqref{relativeL2} and standard deviations, averaged over 10
      independent processes, of the training and testing sets (in
      parentheses) with different $\lambda$ and $\sigma$ when learning
      the dynamics of the noisy data Eq.~\eqref{conv_diff}. An
      increase in $\sigma$ typically results in higher errors.  For
      small $\sigma$, selecting an intermediate $\lambda$ around
      $10^{-1.5}$ balances learning of the adaptive spectral
      parameters (scaling factors $\beta$ and displacements $x_0$)
      with that of coefficients $c_i(t)$ and leads to a minimal
      relative $L^2$ error. For large $\sigma$, choosing a larger
      $\lambda=10^{-0.5}$ to better fit the dynamics of $\beta$ and
      $x_0$ leads to smaller errors.}}
\vskip 0.1in
\small
\renewcommand*{\arraystretch}{1.2}
    \begin{tabular}{l|cccc}
      \hline
\diagbox{$\lambda$}{$\sigma$} &  $0$ & $10^{-4}$ & $10^{-3}$  \\[2pt]
\hline
    $10^{-2}$ &  {$0.0243\pm0.0055$} & {$0.0454\pm0.0693$}  & {$0.0585\pm0.0196$} \\[2pt]
 &  { ($0.0244\pm0.0058$)} & { ($0.0478\pm0.0632$)}  & { ($0.0632\pm0.0207$)} \\[2pt]
\hline
    $10^{-1.75}$ &    {$0.0209\pm0.0068$} & {$0.0469\pm0.0694$}   &{$0.0572\pm0.0278$}\\[2pt]
    &    { ($0.0210\pm0.0069$)} & {($0.0506\pm0.0767$)}  &{ ($0.0602\pm0.0203$)}\\[2pt]
\hline
$10^{-1.5}$  &    {$0.0129\pm0.0017$} & {$0.0477\pm0.0417$} &{$0.1074\pm0.0656$}\\[2pt]
 &   { ($0.0133\pm0.0020$)} & {  ($0.0495\pm0.0499$)}  &{ ($0.1117\pm0.0675$)}\\[2pt]
\hline
$10^{-1.25}$  &    {$0.0115\pm0.0014$} & {$0.0426\pm0.0249$}  &{$0.1178\pm0.0097$}\\[2pt]
 &    { ($0.0116\pm0.0014$)} &{  ($0.0451\pm0.0291$)}  &{ ($0.1231\pm0.1023$)}\\[2pt]
\hline
$10^{-1}$  &    {$0.0103\pm0.0007$} & {$0.0291\pm0.0137$}  &{$0.0626\pm0.0097$}\\[2pt]
  &    { ($0.013\pm0.0006$)} & {  ($0.0292\pm0.0181$)}  &{ ($0.0646\pm0.0111$)}\\[2pt]
\hline
$10^{-0.75}$  &    {$0.0111\pm0.0006$} & {$0.0284\pm0.0005$} &{$0.0672\pm0.0124$}\\[2pt]
  &    { ($0.0131\pm0.0005$)} & {  ($0.0283\pm0.0004$)}  &{ ($0.0667\pm0.0148$)}\\[2pt]
\hline
$10^{-0.5}$  &    {$0.0117\pm0.0005$} & {$0.0272\pm0.0004$}  &{$0.0573\pm0.0009$}\\[2pt]
  &   { ($0.0141\pm0.0004$)} & {  ($0.0271\pm0.0006$)}  &{ ($0.0563\pm0.0006$)}\\[2pt]
\hline
    \end{tabular}%
      \centering
  \label{tab:noise}
\vskip -0.2in
\end{table*}

\section{How noise in the parameter of the solutions affect the learned DE model}
\label{intrinsicnoise}
Here, we take different distributions of the three parameters $\xi_1,
\xi_2, \xi_3$ in Eq.~\eqref{udefinition}. We shall use
$\xi_1\sim\mathcal{N}(3, \tfrac{\sigma^2}{4}),\,\, \xi_2\sim
\frac{1}{4}\mathcal{U}(-\frac{\sigma}{4}, \tfrac{\sigma}{4}),\,\,
\xi_3\sim\mathcal{N}(0, \tfrac{\sigma^2}{2})$ and vary
$\sigma=0,\frac{1}{4}, \frac{1}{2}, 1$.  For different $\sigma$, we
train 10 independent models and the results are as follows.

\begin{table*}[h]
\caption{{\small Training and testing errors of trained models
    on different testing sets with different variances in the initial
    condition parameters, averaged over ten trained models. }}
\small
\vspace{3mm}
\renewcommand*{\arraystretch}{1.2}
    \begin{tabular}{l|cccccc}
\hline
Train $\sigma$ & 0& $\tfrac{1}{4}$
 & \!\!$\tfrac{1}{2}$ & $1$  \!\! \\[2pt]
\hline
  Training error   & 4.43e-03$\pm$9.87e-04 &1.25e-02$\pm$5.34e-04& 1.64e-02$\pm$8.03e-03 & 1.62e-02$\pm$1.10e-03 \\[2pt]
\hline
\diagbox{Train $\sigma$}{Test $\sigma$} & 0& $\tfrac{1}{4}$
& \!\!$\tfrac{1}{2}$ & $1$   \!\! \\[2pt]
\hline
    $0$ & 4.43e-03$\pm$9.87e-04 & 2.81e-02$\pm$3.20e-03 &3.88e-02$\pm$3.26e-03 & 5.64e-02$\pm$3.03e-03  \\[2pt]
    $\tfrac{1}{4}$ &  1.11e-02$\pm$6.55e-04 &  1.28e-02$\pm$5.77e-04 & 2.33e-02$\pm$1.40e-03 & 4.55e-02$\pm$2.40e-03 \\[2pt]
$\tfrac{1}{2}$ & 1.24e-02$\pm$1.025e-03 &  1.07e-02$\pm$6.23e-04 & 1.65e-02$\pm$9.35e-04 & 2.95e-02$\pm$1.45e-03 \\[2pt]
$1$ &1.24e-02$\pm$1.07e-03 &  9.22e-03$\pm$8.66e-04 & 1.03e-02$\pm$7.98e-04 & 1.68e-02$\pm$1.24e-03 \\[2pt]
\hline
    \end{tabular}%
  \centering
  \label{tab:example2_ic}
\end{table*}

The training and testing error is the same for models with $\sigma=0$
in Table~\ref{tab:example2_ic} because if there is no uncertainty in
the initial condition, all trajectories are the same. Though giving
larger training errors, models trained on training sets with larger
variances in the parameters of the initial condition could generalize
better on testing sets where the variances in the parameters of the
initial condition is larger.

\section{Choosing the hyperbolic cross space with different $N$ and $\gamma$}
\label{appendix_gamma}

If the spectral expansion order $N$ is sufficiently large, using a
hyperbolic cross space Eq.~\eqref{hyper_space} can effectively reduce
the required number of basis functions while maintaining accuracy. In
our experiments, we set $N=5, 9, 14$ and $\gamma=-\infty$ (full tensor
product), $-1, 0, \tfrac{1}{2}$. We train the network for 1000 epochs
using SGDM with a learning rate $\eta=0.001$, $\text{momentum}=0.9$,
and $\text{weight decay}=0.005$. The penalty coefficient in
Eq.~\eqref{L2unboundedloss_new} is $\lambda=0.02$.

\begin{table*}[htb]
\vskip -0.15in
    \caption{{\small The $L^2$ errors (Eq.~\eqref{relativeL2}) and the
        total number of basis functions used to learn the evolution of
        the Gaussian wave packet (Eq.~\eqref{wigner_solution}) are
        listed for different $N$ and $\gamma$. The bold number in each
        cell represents the number of basis functions used, which is
        also sensitive to the hyperbolicity $\gamma$. The numbers in
        parentheses are the errors on the testing set.}}
    \small
\vspace{3mm}
\renewcommand*{\arraystretch}{1.2}
    \begin{tabular}{r|ccc}
\hline
\diagbox{$\gamma$}{$N$} & 5 & 9 & $14$ \\[2pt]
\hline
    $-\infty$ &  \textbf{36}, 0.0180 (0.0164)  & \textbf{100}, {0.0204, (0.0194)} & \textbf{225}, {0.0110, (0.0105)}\\[2pt]
    $-1$ &   \textbf{23}, {0.0342}, (0.0302)& \textbf{51}, {0.0185, (0.0166)}& \textbf{92}, {0.0102, (0.0093)}\\[2pt]
$0$  &   \textbf{21}, {0.0414, (0.0363)}& \textbf{42}, {0.0230, (0.0210)}&\textbf{70}, {0.0309, (0.0279)}\\[2pt]
$\frac{1}{2}$  & \textbf{20}, {0.0459, (0.0409)}& \textbf{37}, {0.0413, (0.0365)}
& \textbf{55}, {0.0336, (0.0295)}\\[2pt]
\hline
    \end{tabular}%
    \centering
 \label{tab:wigner}
\end{table*}

\begin{figure}[htb]
\begin{center}
\centerline{\includegraphics[width=\textwidth]{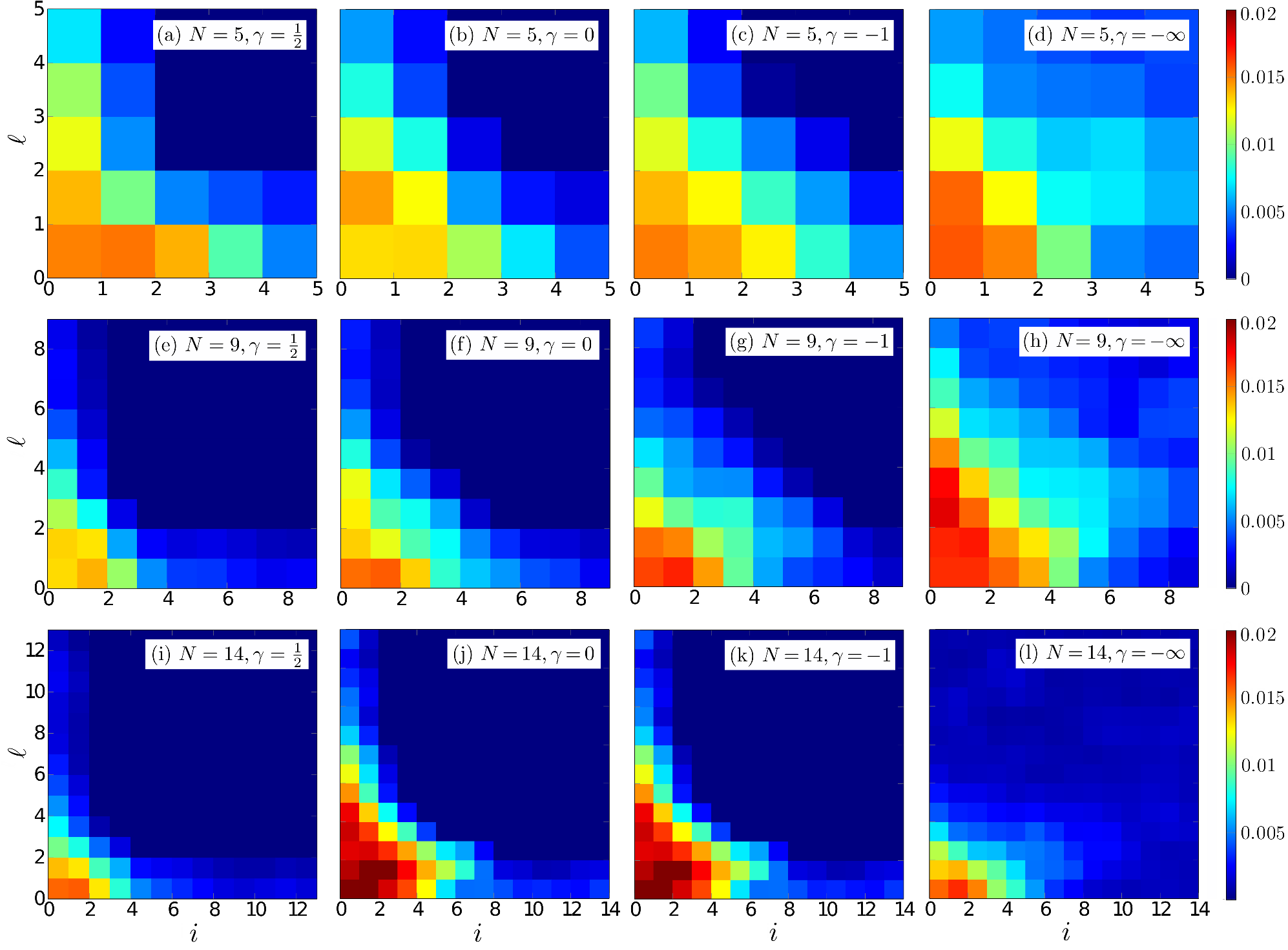}}
\vspace{-0.1in}
\caption{\small Saliency maps of the absolute value of derivatives of
  the relative $L^2$ loss w.r.t. $\{c_{i, \ell}\}$. Here, $N=5, 9, 1$
  and $\gamma=-\infty, -1, 0, \tfrac{1}{2}$ in
  Eq.~\eqref{hyper_space}.  The loss function is always most sensitive
  to frequencies $c_{i, \ell}$ on the lower left of the saliency maps.
  Such a ``resolution-invariance'' indicates that having a larger $N$
  but a moderate $\gamma> -\infty$ to include the frequencies in the
  lower-left part of this saliency map leads to a balance between
  efficiency and accuracy.}
\label{saliency_map}
\end{center}
\vspace{-0.2in}
\end{figure}

From Table~\ref{tab:wigner}, we see that a hyperbolic space with
$N=14, \gamma=-1$ leads to minimal errors on the testing
set. Furthermore, the number of basis functions for the hyperbolic
space with $N=15, \gamma=-1$ is smaller than the full tensor product
space for $N=9, 14$ when $\gamma=-\infty$, so the hyperbolic space
with $N=14, \gamma=-1$ could be close to the most appropriate
choice. We shall also use the saliency map to investigate the role of
different frequencies and plot $|\frac{\partial\text{Loss}}{\partial
  c_{i,\ell}(0)}|$ for different $N$ and $\gamma$ in
Fig.~\eqref{saliency_map}, where the loss function is
Eq.~\eqref{L2unboundedloss_new}. Even for different choices of $N,
\gamma$, the changes in frequencies on the lower-left part of the
saliency map (corresponding to a moderate $\gamma$ and a large $N$)
have the largest impact on the loss. This ``resolution-invariance''
justifies our choices that the proper hyperbolic space should have a
larger $N$ but a moderate $\gamma>\infty$ so that the total number of
inputs or outputs are reduced to boost efficiency in
higher-dimensional problems while accuracy is maintained.

Note that the errors in Table~\ref{tab:wigner} can be larger on the
training set than on the testing set, especially for larger training
set errors (\textit{e.g.}, for $N=5$). This arises because the largest
sampling time among the training samples is $t=1$ while for the
testing samples the largest time is smaller than 1. If the trained
dynamics $F(\tilde{U};t,\Theta)$ does not approximate the true
dynamics $F(\tilde{U};t)$ in Eq.~\eqref{unbounded_dynamics} well, the
error of the training samples with time $t=1$ will be larger than the
error of testing samples due to error accumulation.

\section{Reconstructing learned dynamics}
\label{rec_appendix}
Table~\ref{tab:random_time} shows that the errors from a new testing
set, with randomly sampled time $t_i\sim\mathcal{U}(0, 1.5)$,
$i=1,\ldots,9$, do not differ significantly from the errors in the
first row of Table~\ref{tab:noise}. This suggests that our PDE learner
can accurately extrapolate the solution to times beyond those of the
training samples.
\begin{table}[htb]
\vskip -0.1in
  \caption{{\small Relative $L^2$ errors defined in
      Eq.~\eqref{relativeL2} of trained models with different
      $\lambda$ on a new testing set of samples with noise-free
      coefficients ($\sigma = 0$) and random sampling times $t_j
      \sim\mathcal{U}(0, 1.5), j=1,\ldots,9$.}}
  \centering
\vspace{3mm}
\small
\renewcommand*{\arraystretch}{1.2}
   \begin{tabular}{lcccccccc}
\hline
         {$\lambda$} &  $10^{-2}$ & $10^{-1.75}$ & $10^{-1.5}$ & $10^{-1.25}$ &  $10^{-1}$ & $10^{-0.75}$ & $10^{-0.5}$ 
 \\[2pt] 
 \hline
 {Mean relative $L^{2}$ error} &  $0.1977$ & $0.0944$ &
$0.0229$ & $0.0177$  &  $0.0158$ & $0.0165$ &
$0.0156$  \\[2pt] 
 {s.d. of relative $L^2$ error}  & $0.2590$ & $0.1433$ &
$0.0039$ & $0.0022$  &  $0.0013$ & $0.0012$ &
$0.0006$  \\[2pt] 
\bottomrule
   \end{tabular}
      \centering
  \label{tab:random_time}
\end{table}

To make a comparison with the s-PINN method proposed in
\cite{xia2022spectrally}, we consider the following inverse-type
problem of reconstructing the unknown potential $f(x, t)$ in
\begin{equation}
u_t = u_{xx} + f(x, t),
\label{learn_dynamics}
\end{equation}
by approximating $f(x, t)\approx\hat{f}\coloneqq F[u;x, t, \Theta] -
u_{xx}$. The function $u$ is taken to be
\begin{equation}
u(x, t) = \frac{\xi}{\sqrt{t+1}}\exp(-\tfrac{x^2}{4(t+1)}) + \frac{\sin(x)}{\sqrt{t+1}} \exp(-\tfrac{x^2}{4(t+1)})
\end{equation}
where $\xi\sim\mathcal{U}(\frac{1}{2}, 1)$ is i.i.d. sampled for
different trajectories. Therefore, the true potential in
Eq.~\eqref{learn_dynamics} is
\begin{equation}
f(x, t) = \big[(t+1)\sin(x) + x \cos(x)\big](t+1)^{(-3/2)}\exp(-\tfrac{x^2}{4(t+1)}),
\end{equation}
which is independent of $u(x,t)$. We use 100 trajectories $u(x, t_i)$
to learn the unknown potential with $t_i=i\Delta{t}, \Delta{t}=0.1,
i=0,...,10$. In the s-PINN method, since only $t$ is inputted, only
one reconstructed $\hat{f}$ (which is the same for all trajectories)
is outputted in the form of a spectral expansion. However, in our
spectral neural DE learning method, $f(x, t)\approx\hat{f} = F[u;x, t,
  \Theta] - u_{xx}$ will be different for different inputted $u$
giving rise to a changing error along the time horizon. The mean and
variance of the relative $L^2$ error $\frac{\|\hat{f}-f\|_2}{\|f\|_2}$
is plotted in Fig.~\ref{learn_dynamics}.

\begin{figure}[htb]
\begin{center}
\centerline{\includegraphics[width=3in]{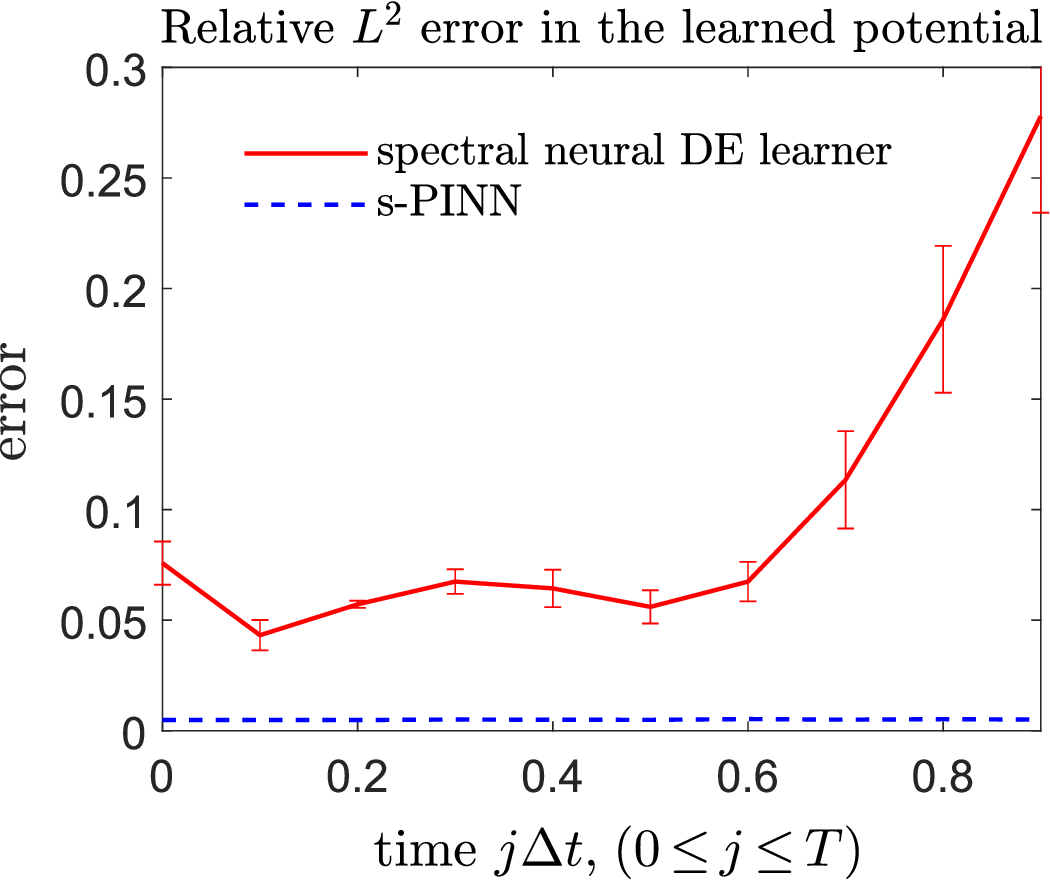}}
\caption{\small Comparison of the relative $L^{2}$ error in the
  potential in Eq.~\eqref{learn_dynamics} learned from our proposed
  spectral neural DE learner and from the s-PINN method. Our spectral
  neural DE learner achieved an average relative $L^{2}$ error of
  $0.1$, while the s-PINN method, designed to input the exact form of
  the RHS of Eq.~\eqref{learn_dynamics} with only one unknown
  potential, achieved better accuracy with an average relative $L^{2}$
  error of about $0.01$.}
\label{learn_dynamics}
\end{center}
\end{figure}

When all but the potential on the RHS of Eq.~\eqref{objective} is
known, the s-PINN is more preferable because more information is
inputted as part of the loss function in
\citep{xia2022spectrally}. Nevertheless, our spectral neural DE
learner can still achieve a relative $L^2$ error $\sim 0.1$,
indicating that without any prior information it can still reconstruct
the unknown source term with acceptable accuracy.  However, the
accuracy of our spectral neural DE learner for reconstructing the
potential $f$ deteriorates as the time horizon $t=j\Delta t$
increases. Since solutions at later times include errors from earlier
times, minimizing Eq.~\eqref{L2unboundedloss_new} requires more
accurate reconstruction of the dynamics (RHS of Eq.~\eqref{objective})
at earlier times than at later times.



\end{document}